\documentclass[11pt]{article}

\usepackage[letterpaper,top=1in,bottom=1in,left=1in,right=1in,
            headheight=14pt,headsep=18pt,footskip=30pt]{geometry}

\usepackage[utf8]{inputenc}
\usepackage[T1]{fontenc}
\usepackage{amsmath}
\usepackage{amsthm}
\usepackage[sups]{XCharter}                 
\usepackage[xcharter,vvarbb]{newtxmath}     
\usepackage[scale=0.95]{sourcesanspro}      
\usepackage[scaled=0.9]{zi4}                
\usepackage[dvipsnames,table]{xcolor}
\definecolor{ucmaroon}{HTML}{800000}        
\definecolor{ucmaroondark}{HTML}{5A0000}
\definecolor{ucgray}{HTML}{767676}          
\definecolor{ucgraylight}{HTML}{D6D6CE}     
\definecolor{ucink}{HTML}{1F1F1F}
\colorlet{ucmaroontint}{ucmaroon!6}
\definecolor{uchicagomaroon}{HTML}{800000}  

\usepackage{url}
\usepackage{multirow}
\usepackage{multicol}
\usepackage{graphicx}
\usepackage{algorithm}
\usepackage{algpseudocode}
\usepackage{array}
\usepackage{makecell}
\usepackage{booktabs}
\usepackage{nicefrac}
\usepackage{microtype}
\usepackage{newfloat}
\usepackage{float}
\usepackage{needspace}
\usepackage{etoc}
\usepackage{fvextra}
\usepackage{wrapfig}
\usepackage{caption}
\usepackage{subcaption}
\usepackage{titlesec}
\usepackage{fancyhdr}
\usepackage{enumitem}
\usepackage[most]{tcolorbox}
\usepackage[round,sort&compress]{natbib}
\usepackage[colorlinks=true,linkcolor=ucmaroon,citecolor=ucmaroon,
            urlcolor=ucmaroon,filecolor=ucmaroon,
            pdfstartview=FitH,bookmarksnumbered=true,linktoc=page]{hyperref}
\usepackage[capitalize]{cleveref}

\usepackage{amsmath,amsfonts,bm}

\def\eqref#1{equation~\ref{#1}}

\def\1{\bm{1}}

\DeclareMathAlphabet{\mathsfit}{\encodingdefault}{\sfdefault}{m}{sl}
\SetMathAlphabet{\mathsfit}{bold}{\encodingdefault}{\sfdefault}{bx}{n}

\titleformat{\section}
  {\sffamily\Large\bfseries\color{ucink}}
  {\color{ucmaroon}\thesection}{0.7em}{}
\titleformat{\subsection}
  {\sffamily\large\bfseries\color{ucink}}
  {\color{ucmaroon}\thesubsection}{0.6em}{}
\titleformat{\subsubsection}
  {\sffamily\normalsize\bfseries\color{ucink}}
  {\color{ucmaroon}\thesubsubsection}{0.6em}{}
\titleformat{\paragraph}[runin]
  {\sffamily\normalsize\bfseries\color{ucink}}{}{0pt}{}[.]
\titlespacing*{\section}{0pt}{2.0ex plus .6ex minus .2ex}{1.0ex plus .2ex}
\titlespacing*{\subsection}{0pt}{1.6ex plus .5ex minus .2ex}{0.7ex plus .2ex}
\titlespacing*{\subsubsection}{0pt}{1.3ex plus .4ex minus .2ex}{0.5ex}

\setlist{itemsep=0.15em,topsep=0.3em,leftmargin=1.4em}
\setlist[itemize]{label={\color{ucmaroon}\small$\blacktriangleright$}}

\DeclareCaptionFont{ucmaroon}{\color{ucmaroon}}
\renewcommand{\bibfont}{\small}
\newtheoremstyle{ucthm}{0.6em}{0.6em}{\itshape}{}
  {\sffamily\bfseries\color{ucmaroon}}{.}{0.5em}{}
\newtheoremstyle{ucdef}{0.6em}{0.6em}{\normalfont}{}
  {\sffamily\bfseries\color{ucmaroon}}{.}{0.5em}{}
\theoremstyle{ucdef}

\theoremstyle{ucthm}

\newtcolorbox{takeaway}[1][Key takeaways]{%
  enhanced, breakable, colback=ucmaroontint, colframe=ucmaroon,
  boxrule=0pt, leftrule=2.5pt, arc=0pt, outer arc=0pt,
  left=10pt, right=10pt, top=6pt, bottom=6pt,
  fonttitle=\sffamily\bfseries\small, coltitle=ucmaroon,
  attach title to upper, title={#1\enspace}}

\makeatletter
\newcommand{\@authorsline}{}
\newcommand{\@affiliationsline}{}
\newcommand{\@metadatalines}{}
\newcommand{\@shorttitle}{}
\newcommand{\@toplabel}{Preprint}
\newcommand{\authors}[1]{\renewcommand{\@authorsline}{#1}}
\newcommand{\affiliations}[1]{\renewcommand{\@affiliationsline}{#1}}
\newcommand{\shorttitle}[1]{\renewcommand{\@shorttitle}{#1}}
\newcommand{\toplabel}[1]{\renewcommand{\@toplabel}{#1}}
\newcommand{\metadata}[2]{%
  \g@addto@macro\@metadatalines{%
    \par\noindent{\sffamily\bfseries\color{ucink}#1:}\enspace{\sffamily #2}}}
\newcommand{\affilmark}[1]{\textsuperscript{\color{ucmaroon}#1}}
\newcommand{\email}[1]{\href{mailto:#1}{\nolinkurl{#1}}}

\renewcommand{\maketitle}{%
  \thispagestyle{firstpage}%
  \begingroup
  \setlength{\parskip}{0pt}%
  \noindent{\sffamily\small\bfseries\color{ucmaroon}\MakeUppercase{\@toplabel}}\par
  \vspace{6pt}{\color{ucmaroon}\rule{\linewidth}{1.6pt}}\par
  \vspace{20pt}
  {\raggedright\sffamily\bfseries\fontsize{23}{28}\selectfont\color{ucink}
   \@title\par}
  \vspace{14pt}
  {\raggedright\sffamily\large\color{ucink}\@authorsline\par}
  \ifx\@affiliationsline\@empty\else
    \vspace{6pt}
    {\raggedright\sffamily\small\color{ucgray}\setlength{\parskip}{2pt}\@affiliationsline\par}
  \fi
  \vspace{16pt}
  \endgroup
}

\renewenvironment{abstract}{%
  \par\noindent\begingroup\setlength{\parskip}{0.5em}%
  {\sffamily\bfseries\small\color{ucmaroon}\MakeUppercase{Abstract}}\par
  \vspace{2pt}%
}{%
  \par
  \ifx\@metadatalines\@empty\else
    \vspace{10pt}
    {\small\setlength{\parskip}{3pt}\@metadatalines\par}
  \fi
  \vspace{12pt}{\color{ucgraylight}\rule{\linewidth}{0.6pt}}\par
  \endgroup
  \vspace{6pt}%
}

\fancypagestyle{firstpage}{%
  \fancyhf{}%
  \renewcommand{\headrule}{}%
  \fancyfoot[C]{\sffamily\small\color{ucgray}\thepage}}
\fancypagestyle{main}{%
  \fancyhf{}%
  \fancyhead[L]{\sffamily\footnotesize\color{ucgray}\@shorttitle}%
  \fancyhead[R]{\sffamily\footnotesize\bfseries\color{ucmaroon}\thepage}%
  \renewcommand{\headrule}{{\color{ucgraylight}%
    \hrule width\headwidth height 0.4pt \vskip-0.4pt}}}
\newcommand{\appendixcontents}{%
  \begingroup
  \setlength{\parskip}{0pt}%
  \noindent{\color{ucmaroon}\rule{\linewidth}{1.6pt}}\par\vspace{8pt}
  {\sffamily\bfseries\fontsize{20}{24}\selectfont\color{ucink}Appendix\par}
  \vspace{4pt}
  \etocsettagdepth{mainmatter}{none}%
  \etocsettagdepth{appendix}{subsection}%
  \etocsettocstyle{\vspace{4pt}{\sffamily\small\bfseries\color{ucmaroon}%
    \MakeUppercase{Contents}}\par\vspace{2pt}}{\vspace{6pt}%
    {\color{ucgraylight}\rule{\linewidth}{0.6pt}}\par}%
  \etocsetstyle{section}{}{}%
    {\par\vspace{6pt}\noindent{\sffamily\bfseries
      \makebox[1.8em][l]{\color{ucmaroon}\etocnumber}%
      \etocname\leaders\hbox to .6em{\hss.\hss}\hfill\etocpage}\par}{}%
  \etocsetstyle{subsection}{}{}%
    {\par\vspace{1pt}\noindent\hspace{1.8em}{\small
      \makebox[2.4em][l]{\color{ucgray}\etocnumber}%
      \etocname\leaders\hbox to .6em{\hss.\hss}\hfill\etocpage}\par}{}%
  \tableofcontents
  \endgroup
  \vspace{10pt}
}
\makeatother



\newif\ifshowcomments
\showcommentsfalse
\ifshowcomments
  \newcommand\dixi[1]{{\color{blue}\textbf{DY: }{#1} }}
  \newcommand\xc[1]{{\color{cyan}\textbf{XC: }{#1} }}
  \newcommand\yf[1]{{\color{purple}\textbf{YF: }{#1} }}
  \newcommand{\chenhao}[1]{{\color{magenta}\textbf{CT: }{#1} }}
  \newcommand{\shane}[1]{{\color{uchicagomaroon}\textbf{SG: }#1}}
\else
  \newcommand\dixi[1]{}
  \newcommand\xc[1]{}
  \newcommand\yf[1]{}
  \newcommand{\chenhao}[1]{}
  \newcommand{\shane}[1]{}
\fi

\title{TIGER: Learning from Researcher Taste to Generate Tailored Research Ideas}
\shorttitle{TIGER: Learning from Researcher Taste to Generate Tailored Research Ideas}
\toplabel{arXiv preprint}

\authors{%
  Shane Gu\affilmark{1,*},\enspace
  Yufeng Lin\affilmark{2,*},\enspace
  Dixi Yao\affilmark{1,*},\enspace
  Xuchen Gong\affilmark{1,*},\enspace
  Chenhao Tan\affilmark{1}}
\affiliations{%
  \affilmark{1}\,University of Chicago \par
  \affilmark{2}\,The Chinese University of Hong Kong, Shenzhen \par
  \affilmark{*}\,Equal contribution}
\metadata{Correspondence}{\email{dixi@uchicago.edu}}

\begin{document}
\etocdepthtag.toc{mainmatter}
\raggedbottom

\maketitle

\begin{abstract}
Large language models (LLMs) can generate research ideas, but their outputs often fail to match an individual researcher's taste. We introduce Tailored Idea GEnerator for Research (TIGER), a framework that learns individual preferences to guide idea generation. Using 400 real-world research requests and 2,000 candidate ideas, we collect ratings from three LLM researchers and two PhD researchers. For each researcher, we train a separate Qwen3.8-27B reward model (RM) on their annotations, then fix its parameters and use its scores to optimize a Qwen3-8B ideation policy through reinforcement learning. Neutral rewriting and auxiliary quality checks guide optimization alongside the preference reward.  
In pairwise comparisons where each judge chooses between an idea generated by its corresponding TIGER-8B policy and one generated by GPT-4.1, the three LLM researchers select the TIGER-8B idea in 56.6\%-68.4\% of 76 comparisons per researcher. PhD~1 and PhD~2 select the TIGER-8B idea in 19 of 30 pairs (63.3\%) and 17 of 30 pairs (56.7\%), respectively. All six directed cross-policy comparisons likewise favor the judge's matched TIGER-8B policy over an off-target policy, supporting the use of individual preference feedback for tailored research ideation.
\end{abstract}
\section{Introduction}
Recent work uses large language models (LLMs) to generate research ideas and hypotheses from scientific literature and observed data \citep{scimon,hypogenic,literature_meets_data,baek2025researchagent,lu2024ai,yamada2025ai,yu2024researchtown}. Co-Scientist iteratively generates, critiques, and refines hypotheses \citep{co_scientist}, while NeuriCo connects literature review with experiment execution and analysis \citep{neurico}.
These advances expand the research tasks that AI systems can support, but do not by themselves ensure that the proposed ideas reflect an individual researcher's priorities. Choosing which research directions to pursue involves \textit{research taste}, a form of judgment discussed in human research and increasingly examined in LLMs \citep{olah2021taste,tastybench,scientific_taste}.

Rather than imposing a single definition of a ``good'' research idea, our objective is to generate ideas that better match the target researcher's preferences, conditioned on a research request and literature context. The researcher may be a human researcher or an LLM; this formulation does not assume that their preferences coincide. We therefore ask: \textit{Can a target researcher's research taste be learned from a fixed set of preference judgments and used to guide the generation of new research ideas?}

\begin{figure}[!t]
    \centering
    \includegraphics[width=\linewidth]{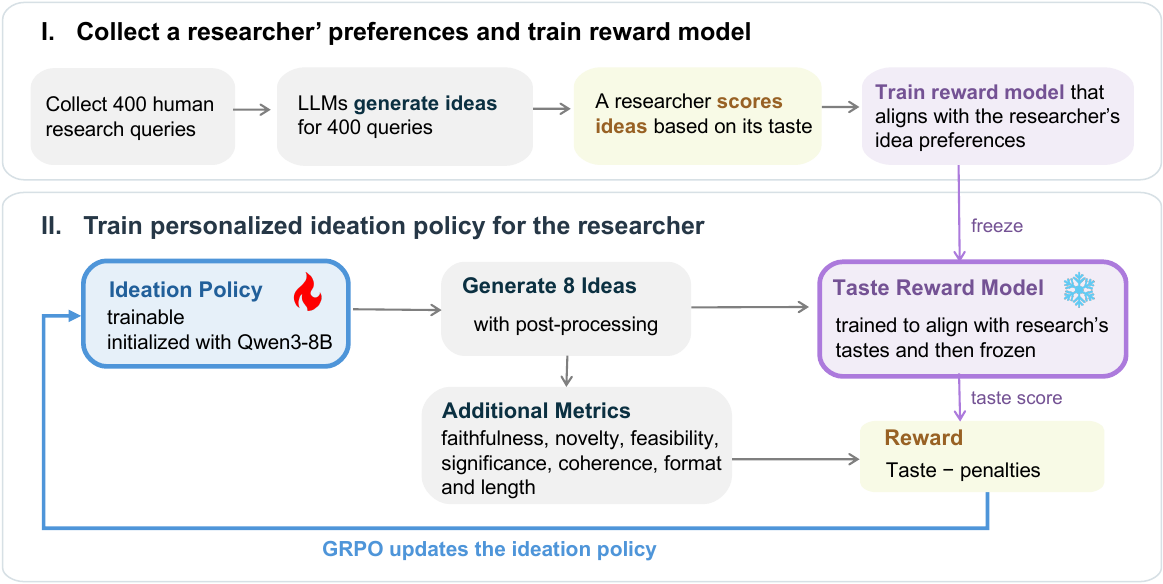}
    \caption{TIGER training pipeline, repeated for each researcher. (I) A researcher's ratings of candidate ideas generated from research requests train a reward model. (II) A policy initialized from Qwen3-8B samples eight ideas per request. Post-processing uses a frozen neutral rewriter to supply the RM input; a frozen quality judge scores five quality dimensions, with format and length checked separately. The taste score and penalties form the reward in Eq.~\ref{eq:ideation_policy_reward}, which drives GRPO updates on original-response tokens. Only the policy is updated; the RM, rewriter, and quality judge remain frozen.}
    \label{fig:framework-detail}
\end{figure}

First, a major challenge is that collecting preference data for research ideation requires examples grounded in a concrete research context. Each example must pair a research request and its relevant literature with alternative ideas that the target researcher can meaningfully compare. Evaluating these alternatives requires understanding both the proposed contributions and their relation to prior work, making human annotation time-consuming. A practical approach must therefore learn effectively from a finite set of judgments that a researcher can reasonably provide.

To address this challenge, we propose TIGER (\textbf{T}ailored \textbf{I}dea \textbf{GE}nerator for \textbf{R}esearch), a framework that learns a researcher's taste from offline judgments and uses it to train a compact ideation policy. Figure~\ref{fig:framework-detail} illustrates its two training stages. For research personalization, a key principle is that the judgments of a general-purpose LLM should not be assumed to reflect the preferences of a particular researcher~\citep{dong2024can,chen2025pal}. TIGER therefore learns a researcher-specific reward model (RM) that captures these preferences and generalizes them to unseen query-idea pairs for downstream policy training. Researchers may also prefer to keep unpublished ideas private rather than send them to an external API, further motivating a compact ideation model that can run locally while reflecting their research taste (e.g. an NVIDIA A100).

Therefore, first, we train a separate Qwen3.8-27B RM for each researcher using their annotations to predict both ratings and preference rankings. Second, we freeze the trained RM and use its scores as rewards to optimize a Qwen3-8B policy through reinforcement learning (RL). Thus, the RM learns the target researcher's taste through supervised training, while the policy learns to generate ideas that receive higher personalized rewards. In an initial study within our lab, two PhD students rated a pool of 227 candidate ideas spanning 40 research requests. Their agreement with LLM judges varied substantially across researchers (Appendix~\ref{app:exploratory_agreement}, Figure~\ref{fig:taste-agreement}), further motivating researcher-specific modeling rather than reliance on a single general-purpose judge.


Our contributions are as follows:

1) A benchmark for personalized research ideation, comprising 400 real-world research requests and 2,000 candidate ideas generated by five LLM backbones. The benchmark includes preference annotations from three LLM researchers and two PhD researchers, each providing ratings over 2,000 ideas, enabling systematic study of researcher-specific taste.

2) A researcher-specific reward-modeling and policy-optimization framework that converts a finite set of offline preference judgments into a compact personalized ideation model. For each researcher, we train a Qwen3.8-27B reward model on within-request rankings and numerical ratings, then freeze it to optimize a Qwen3-8B policy model with predicted rewards, neutral rewriting, and auxiliary quality penalties, requiring no additional researcher annotation during policy training.

3) An evaluation protocol for tailored research ideation that combines comparisons against GPT-4.1 with cross-policy judgments to test both overall quality and researcher-specific alignment. Under this evaluation, the corresponding TIGER-8B policies achieve win-rate point estimates above 50\% for all three LLM researchers, while both PhD researchers prefer their corresponding policies in a majority of direct comparisons with GPT-4.1. Cross-policy judgments further show that each LLM researcher favors its matched policy over policies trained for the other researchers.

\section{Related Work}
\subsection{Autonomous Researcher}
Automatic idea generation is becoming an important first step toward AI Researchers and has attracted increasing attention. Existing work grounds hypothesis generation in research literature or observed data \citep{scimon,literature_meets_data,hypogenic}. More comprehensive systems, such as AI Co-Scientist \citep{co_scientist} and NeuriCo \citep{neurico}, extend this process to multiple stages of research, from literature review to experimental design. To improve research quality, GIANTS \citep{giants} predicts core insights by combining two parent papers, while EvoIdeator \citep{evoideator} uses checklist-grounded feedback in a reinforcement learning pipeline. However, existing work primarily focuses on generating research ideas, while paying limited attention to judging whether the generated ideas align with a particular researcher's preferences.

\subsection{AI Research Taste}
Research taste has been studied through citation-based measures \citep{measure_research_taste} and institutional publication records \citep{institutional_taste}. AI Can Learn Scientific Taste \citep{scientific_taste} learns a scientific judge from citation-based community preferences and uses it as a reward model to train an idea generator with reinforcement learning. TIGER instead learns a separate reward model from each researcher's own ratings of candidate ideas and uses it to align the corresponding ideation policy with that researcher's preferences. Personalized Auto-Research \citep{personalized_autoresearch} further proposes conditioning research workflows on a researcher's prior work, methods, and collaborators. LLM-generated ideas also differ systematically from human research ideas in how they frame opportunities and construct contributions \citep{human_llm_ideas_gap}. Benchmarks evaluate both hypothesis generation \citep{hypobench} and the alignment of idea evaluations with expert judgments \citep{ligbench,ideation_arena}. However, human studies and expert evaluations still identify limitations in LLM self-evaluation and agreement with expert preferences \citep{si2025ideas,ideation_arena}.

\subsection{Preference Alignment}
Preference learning uses human or AI feedback to guide model behavior \citep{christiano2017preferences,rlaif}, with personalized approaches further accounting for differences between individual users \citep{prlhf,pal,reward_features}. These methods provide a basis for learning research taste. Our setting requires learning a particular researcher's preferences over research ideas from a fixed set of judgments and using them to guide new generations, without collecting further judgments from that researcher during policy training.

Reward design also depends on the task. Mathematical reasoning and code generation can use correctness checks \citep{process_reward,code_reward}, while agent reward models assess progress toward a task goal \citep{agentprm}. Research ideas require judgments about their proposed contributions and relation to existing literature, without a single correctness criterion that captures a researcher's preferences. Preference models and task-specific rubrics for open-ended generation \citep{writing_zero,rubric_generation} provide ways to learn such reward signals.

Finally, optimizing an imperfect reward model can increase predicted scores without improving the outputs according to the target researcher \citep{reward_overoptimization,reward_ensembles}. In research ideation, a policy may receive high rewards for polished descriptions, repackaged literature, or unsupported claims of completed experiments. We therefore combine preference rewards with neutral rewriting and auxiliary quality checks.

\section{Tailored Idea GEnerator for Research (TIGER)}
TIGER first learns a reward model from a researcher's annotations and then uses the trained model to optimize an ideation policy (Figure~\ref{fig:framework-detail}). We describe data collection, supervised RM training, and policy optimization in that order.

\subsection{Data Collection}
\label{sec:data_collection}
We collect 400 real-world research requests submitted to Hypogenic AI. We use the same ideation agent with five LLM backbones: Qwen3-8B, DeepSeek V4.1 Flash, Qwen3-32B, Llama 3.3-70B Instruct, and GPT-5.6 Sol. The agent generates ideas from each request and its literature review. We retain 2,000 ideas, with 400 contributed by each backbone and five candidates per request. Before evaluation, we group candidates by request and hide their model identities and generation order.

\begin{figure}[t]
    \centering
    \includegraphics[width=\linewidth]{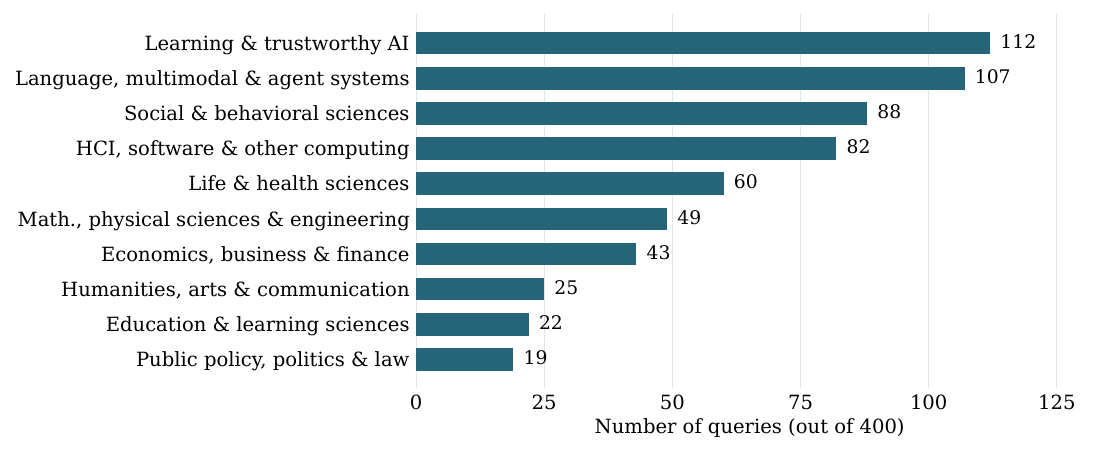}
    \caption{Research areas covered by the 400 requests, using LLM-assisted labels. Queries may belong to multiple areas. HCI denotes human--computer interaction.}
    \label{fig:query-domains}
\end{figure}

The requests span learning and trustworthy AI; language, multimodal, and agent systems; social and behavioral sciences; HCI, software, and other computing; life and health sciences; mathematics, physical sciences, and engineering; economics, business, and finance; humanities, arts, and communication; education and learning sciences; and public policy, politics, and law. Figure~\ref{fig:query-domains} shows the coverage of these ten categories; a request may belong to multiple categories.

The three LLM researchers and two PhD researchers each rate 2,000 ideas for the same 400 requests, and their reward models use these annotations. In the human interface, researchers see a research query, its literature context, and several candidate ideas. They rate each idea on a five-point preference scale from [-2, -1, 0, 1, 2], indicating how much they like or dislike it. These ratings capture each researcher's overall preference for an idea. We retain each researcher's ratings separately to study individual preferences. These full-pool annotations are distinct from the pilot ratings in \Cref{fig:human-llm-taste}.

LLM researchers follow a shared evaluation protocol with the same scoring dimensions, score scales, and output format. We retain each researcher's component scores and sum them to obtain its overall score for an idea. The researchers, scoring rubric, and prompts used are described in \Cref{sec:rm_evaluation}.

We operationalize research taste through a researcher's relative preferences among candidate ideas proposed for the same query. For human researchers, these preferences are obtained from overall ratings; for LLM researchers, they are obtained from summed component scores. Using the same candidate pool and evaluation protocol across LLMs helps separate differences in their preferences from differences in task presentation. The exploratory study in \Cref{fig:taste-agreement}(a) shows that agreement varies across LLM pairs, with within-query Spearman correlations ranging from 0.51 to 0.62.

\subsection{Learning Researcher-Specific Reward Models}
\label{sec:reward_model}

We initialize a separate reward model from Qwen3.8-27B for each researcher and train it on that researcher's offline annotations. The model learns to predict the researcher's numerical ratings and preference rankings, yielding a scalar taste reward for a request--idea pair. For example, the RM trained on Haiku's ratings is a Qwen3.8-27B model supervised by annotations from Haiku. Researcher names identify the source of preference supervision; all five RMs share the Qwen3.8-27B backbone. We call a Qwen3-8B policy optimized with this RM the \emph{Haiku-aligned policy}, and use the same convention for the other researchers.

\textbf{Supervision.}
For each query $q_i$, researcher $e$ scores a candidate set $\mathcal{I}_i=\{I_{i,1},\ldots,I_{i,K_i}\}$. Here, $q_i$ includes the literature context when it is supplied to the reward model. The annotation $\mathbf{s}_{i,j}^e$ contains one overall rating or $D$ component scores for idea $I_{i,j}$. We keep each researcher's labels separate and apply a scaling factor $g_e$ for scores in training. The model $f_{\theta_e}$ predicts the corresponding scores and sums them to a scalar reward. An overall rating uses a prediction head, corresponding to $D=1$. Here, $N$ denotes the number of training queries and $B$ denotes a minibatch of queries.

\textbf{Training objective.}
A listwise soft-label cross-entropy loss learns relative preferences among ideas for the same query. Equal total ratings receive equal target probabilities. A total-score Huber loss fits the researcher's numerical scores. When component labels are available, we additionally fit them with a component-level Huber loss. For overall ratings, this component loss is omitted by setting $\lambda_{\mathrm{dim}}=0$. Algorithm~\ref{alg:ideation_rm} gives the shared procedure; the score scaling, prediction head, temperature, and loss weights follow the annotation format and experimental setting.

\begin{algorithm}[t]
\small
\caption{Training a Researcher-Specific Reward Model}
\label{alg:ideation_rm}
\begin{algorithmic}[1]
\Require Labels $\mathcal{D}_e=\{(q_i,\mathcal{I}_i,\mathbf{S}_i^e)\}_{i=1}^{N}$; $D$-output reward model $f_{\theta_e}$
\Require Score scaling $g_e$; temperature $T>0$; weights $\lambda_{\mathrm{total}},\lambda_{\mathrm{dim}}\geq 0$
\For{each query batch $B\subset\mathcal{D}_e$}
    \For{each query $i$ in $B$}
        \State $\mathbf{y}_{i,j}^{e}\gets g_e(\mathbf{s}_{i,j}^{e})$, $\hat{\mathbf{y}}_{i,j}\gets f_{\theta_e}(q_i,I_{i,j})$ for all $j$
        \State $y_{i,j}^{e}\gets\sum_{d=1}^{D}y_{i,j,d}^{e}$, $r_{i,j}\gets\sum_{d=1}^{D}\hat y_{i,j,d}$ for all $j$
        \State $p_i^e\gets\operatorname{softmax}\big((y_{i,1}^{e},\ldots,y_{i,K_i}^{e})/T\big)$
        \State $\hat p_i\gets\operatorname{softmax}\big((r_{i,1},\ldots,r_{i,K_i})/T\big)$
        \State $\mathcal{L}_{\mathrm{rank}}^{(i)}\gets-\sum_{j=1}^{K_i}p_{i,j}^{e}\log\hat p_{i,j}$
        \State $\mathcal{L}_{\mathrm{total}}^{(i)}\gets\frac{1}{K_i}\sum_{j=1}^{K_i}\ell_{\mathrm{Huber}}(r_{i,j},y_{i,j}^{e})$
        \State $\mathcal{L}_{\mathrm{dim}}^{(i)}\gets\frac{1}{DK_i}\sum_{j=1}^{K_i}\sum_{d=1}^{D}\ell_{\mathrm{Huber}}(\hat y_{i,j,d},y_{i,j,d}^{e})$
        \State $\mathcal{L}_i\gets\mathcal{L}_{\mathrm{rank}}^{(i)}+\lambda_{\mathrm{total}}\mathcal{L}_{\mathrm{total}}^{(i)}+\lambda_{\mathrm{dim}}\mathcal{L}_{\mathrm{dim}}^{(i)}$
    \EndFor
    \State Update $\theta_e$ to minimize $|B|^{-1}\sum_{i\in B}\mathcal{L}_i$
\EndFor
\State \Return $f_{\theta_e}$
\end{algorithmic}
\end{algorithm}

Each query's candidates are kept together during training, and data splits separate entire queries. After supervised training and checkpoint selection, we freeze the RM: its learned parameters remain fixed while it scores new ideas during policy optimization. Personalization is therefore learned before freezing, and subsequent RL updates only the ideation policy. Section~\ref{sec:rm_evaluation} describes the experimental setup, and Appendix~\ref{app:rm_training} provides the model heads, label encodings, and training configurations.

\subsection{Optimizing the Ideation Policy}

For each target researcher, we initialize a separate ideation policy from Qwen3-8B and optimize it using the RM trained on that researcher's annotations. Given a research request $q_i$ and its literature review $L_i$, the policy samples a group of candidate ideas $\{I_{i,j}\}_{j=1}^{K}$. Each candidate receives a scalar reward, and the policy is optimized from rewards relative to the other candidates generated for the same request. The trained RM supplies taste feedback with fixed parameters, without further judgments from the researcher.

\textbf{Rollout feedback.}
A frozen rewriting model converts each candidate $I_{i,j}$ into a neutrally worded version $\widetilde I_{i,j}$ while preserving its scientific content. The researcher-specific reward model assigns a taste score to the rewritten idea, using the same input context as during RM training. Separately, a frozen quality judge evaluates the candidate with $(q_i,L_i)$ for faithfulness, novelty, feasibility, significance, and coherence. We also apply penalties for invalid output format and excessive length. The resulting reward is
\begin{equation}
    R = s_{\mathrm{taste}}
    - p_{\mathrm{faith}} - p_{\mathrm{nov}} - p_{\mathrm{feas}}
    - p_{\mathrm{sig}} - p_{\mathrm{coh}} - p_{\mathrm{fmt}} - p_{\mathrm{len}}.
    \label{eq:ideation_policy_reward}
\end{equation}

\textbf{Reward components.}
The taste reward $s_{\mathrm{taste}}$ measures how well an idea matches the target researcher's preferences. It is computed from the neutrally rewritten idea so that persuasive wording alone does not increase the reward.

The quality penalties target specific weaknesses in a proposed research idea. The faithfulness penalty $p_{\mathrm{faith}}$ applies when the response invents unsupported research gaps, misrepresents prior work, or states expected outcomes as established results. The novelty penalty $p_{\mathrm{nov}}$ applies when the idea largely repeats or repackages existing work. The feasibility penalty $p_{\mathrm{feas}}$ applies when the proposed study is difficult to execute or evaluate. The significance penalty $p_{\mathrm{sig}}$ applies when the idea would offer limited value even if successful. The coherence penalty $p_{\mathrm{coh}}$ applies when the research question, claimed contribution, methodology, and evaluation plan are not well aligned.

The format penalty $p_{\mathrm{fmt}}$ is applied when the response does not follow the required structure, while the length penalty $p_{\mathrm{len}}$ discourages the policy from obtaining a higher reward simply by generating a longer response. Format and length penalties are computed from the sampled response. Section~\ref{sec:policy_evaluation} gives the exact penalty values and training settings.

\textbf{Training objective.}
We optimize the policy with group relative policy optimization (GRPO). For each request, rewards are normalized within its candidate group to compute relative advantages. These advantages weight a PPO-style clipped objective over the tokens of the original sampled responses. Groups with identical rewards do not contribute to the update. The reward model, rewriting model, and quality judge remain fixed throughout policy training. 

We also consider variants without the taste reward, without the frozen quality judge, and without neutral rewriting. These ablations test the roles of researcher-specific taste, auxiliary judging, and presentation style, respectively; Section~\ref{sec:ablation} gives the full settings and results.

\section{Evaluation}
Our evaluation asks whether TIGER can learn a researcher's preferences and use them to generate ideas that the researcher prefers. We compare matched and off-target reward models on validation ratings and evaluate generated ideas against GPT-4.1. Our cross-policy evaluation further examines whether each LLM researcher prefers their matched policy. Ablations examine the roles of the taste reward, neutral rewriting, and auxiliary judging.

\Needspace{6\baselineskip}
\subsection{Reward Model}
\label{sec:rm_evaluation}

\textbf{Researchers and annotation prompts.}
The human setting uses individual preference ratings from two PhD researchers, denoted PhD~1 and PhD~2, each on 2,000 ideas as described in Section~\ref{sec:data_collection}. The three LLM researchers are GPT-5.6 Terra, Gemini 3.7 Flash, and Claude Haiku 4.5. Each LLM scores novelty, significance, plausibility, feasibility, expected effectiveness, and excitement. The prompts request integer scores from 0 to 100, permit ties, and ask researchers to assess the contribution rather than persuasive wording. Novelty is scored both closed-book and with literature, while expected effectiveness uses literature; the remaining dimensions are scored closed-book. This yields seven scores per idea, with a total ranging from 0 to 700.

\textbf{Experimental setup.}
We train a separate Qwen3.8-27B reward model for each researcher using 320 training queries and 80 validation queries. RMs trained on human ratings receive the request and idea, while those trained on LLM ratings also receive the literature review. Appendix~\ref{app:rm_training} provides the model heads, label encodings, optimization settings, checkpoint selection, and score calibration.

\textbf{Comparison and metrics.}
We report pooled strict pair accuracy: the number of correctly ordered idea pairs divided by the number of eligible pairs across queries. An eligible pair has the same strict order in every reference ordering; a predicted tie counts as incorrect. The human labels treat identical ratings as ties. For LLM labels, we sort scores in descending order and form tie groups within five points of each group's highest score on the original 0--700 scale.

To test whether a reward model learns individual preferences, we compare the matched RM, trained on the target researcher's ratings, with off-target RMs trained on other researchers' ratings. For each target researcher, all RMs score the same validation candidates and are evaluated against that researcher's ordering. Table~\ref{tab:rm_cross_eval} reports each comparison separately. We estimate matched-minus-off-target differences with 10,000 paired bootstrap resamples of queries. All models use the same 80 validation queries. Each RM retains its native input format: those trained on LLM ratings receive the literature review, while those trained on human ratings receive only the request and idea. Comparisons within each group share the input format and training candidate pool; comparisons across groups also reflect these differences.

\textbf{Results.}
The matched RM achieves the highest strict pair accuracy for each of the five target researchers. On Haiku's labels, the RM trained on Haiku's ratings outperforms the RM trained on Terra's ratings by 9.03 percentage points (95\% CI: 5.58-12.52). On Gemini's labels, the matched RM outperforms the RM trained on Terra by 8.35 percentage points (95\% CI: 5.63-11.15).

\begin{table}[t]
    \centering
    \small
    \setlength{\tabcolsep}{4pt}
    \setlength{\abovecaptionskip}{4pt}
    \setlength{\belowcaptionskip}{3pt}
    \captionsetup{skip=2pt}
    \caption{Strict pair accuracy (\%) of matched and off-target RMs, all initialized from Qwen3.8-27B. Rows identify the researcher whose labels are used for evaluation; columns identify whose annotations trained the RM. Bold diagonal entries are matched RMs.}
    \label{tab:rm_cross_eval}
    \begin{tabular}{@{}lrrrrr@{}}
        \toprule
        & \multicolumn{5}{c}{Source of RM training annotations} \\
        \cmidrule(l){2-6}
        Target researcher & Terra & Gemini 3.7 & Haiku & PhD~1 & PhD~2 \\
        \midrule
        Terra & \textbf{77.35} & 76.01 & 76.81 & 67.29 & 59.92 \\
        Gemini 3.7 & 83.55 & \textbf{91.90} & 91.26 & 71.08 & 60.28 \\
        Haiku & 82.71 & 90.71 & \textbf{91.74} & 71.74 & 60.00 \\
        \midrule
        PhD~1 & 67.11 & 60.49 & 54.44 & \textbf{76.56} & 74.67 \\
        PhD~2 & 61.96 & 52.97 & 45.81 & 74.64 & \textbf{79.96} \\
        \bottomrule
    \end{tabular}
\end{table}

\Needspace{6\baselineskip}
\subsection{Policy Training and Evaluation}
\label{sec:policy_evaluation}

\textbf{Training settings.}
For each target researcher, we initialize the policy from Qwen3-8B with thinking disabled and train a separate LoRA adapter on all linear modules using the matched reward model. Each rollout generates one idea as a JSON array containing its title, description, and summary (\texttt{tldr}), with a target length of at most 2,000 characters. Optimization settings, including the configuration-specific LoRA scaling, are listed in Table~\ref{tab:policy_settings} in Appendix~\ref{app:policy_optimization}.

\textbf{Data and auxiliary models.}
For policy training, we split the 400 requests into 324 training queries and 76 held-out test queries. Policy optimization uses only the 324 training queries; the remaining 76 queries are reserved for the final pairwise evaluation reported below. Across all policy-training runs, both the neutral rewriter and quality judge are frozen Qwen3.8-27B models. The policy prompt requests a concrete literature-grounded contribution, its distinction from the closest work, and a credible evaluation plan, while keeping expected outcomes hypothetical. The rewriter removes promotional wording and unsupported certainty while preserving scientific content, and the quality judge scores faithfulness, novelty, feasibility, significance, and coherence from 0 to 4 in a single evaluation.

\textbf{Penalty settings.}
In this auxiliary-quality configuration, a frozen quality judge scores faithfulness, novelty, feasibility, significance, and coherence. It returns an integer score $s_d\in\{0,1,2,3,4\}$ for each dimension. We convert it to a penalty $p_d=(4-s_d)/4$, so each dimension contributes between zero and one. An invalid output format incurs a penalty of 2. The length penalty is zero up to 2,000 characters, increases linearly to 2 at 4,000 characters, and remains capped at 2. Format and length checks use the sampled response. The taste score uses the rewritten idea and the frozen reward model, with score calibration described in Appendix~\ref{app:rm_training}.

\textbf{Preference scores.}
For each of the three LLM-aligned policies, we compare the mean taste score of ideas sampled in the initial training batch with that of ideas generated by the corresponding trained TIGER-8B policy, using the same researcher-specific RM for both. The mean score increases from 4.00 to 4.57 for Terra, from 3.64 to 4.77 for Haiku, and from 3.75 to 4.85 for Gemini, corresponding to increases of 0.57, 1.13, and 1.10, respectively. Each RM is held fixed throughout the corresponding policy's training and scoring. All RM scores are measured before subtracting quality, format, and length penalties.

\textbf{Qualitative example.}
Figure~\ref{fig:haiku_qualitative} reproduces the titles and summaries of ideas generated before and after RL on the same request. The later idea specifies a comparison between explicit and implicit state-transition rules, with robustness under perturbations as an evaluation target. Appendix~\ref{app:qualitative_examples} provides the complete original descriptions.

\begin{figure}[t]
    \centering
    \small
    \setlength{\fboxsep}{7pt}
    \fbox{%
    \begin{minipage}{\dimexpr\linewidth-2\fboxsep-2\fboxrule\relax}
        \textbf{Request:} \emph{generate ideas on synthesizing LLM tool learning environments}
        \par\medskip
        \begin{minipage}[t]{0.47\linewidth}
            \raggedright
            \textbf{Qwen3-8B (before RL)}\par
            \textbf{RM score: 3.27}\par\smallskip
            \textit{Synthesizing Tool-Use Environments with Dynamic State and Multi-Agent Interaction}\par\smallskip
            Develop dynamic, multi-agent tool-use environments to enhance agent training through evolving state and interaction complexity.
        \end{minipage}\hfill
        \begin{minipage}[t]{0.47\linewidth}
            \raggedright
            \textbf{Haiku-aligned TIGER-8B}\par
            \textbf{RM score: 4.64}\par\smallskip
            \textit{Evaluating the Role of State Transition Fidelity in Tool-Use Environment Synthesis}\par\smallskip
            Compare agent performance on tool-use tasks using environments with explicit vs. implicit state transition rules, testing whether fidelity to programmatic state validity reduces adversarial/synthetic perturbation vulnerability.
        \end{minipage}
    \end{minipage}%
    }
    \caption{Verbatim titles and summaries from model outputs for the same request, literature review, and generation prompt. The RM trained on Haiku's ratings scores neutral rewrites of the complete ideas at 3.27 and 4.64, before penalties. Appendix~\ref{app:qualitative_examples} reproduces the full descriptions.}
    \label{fig:haiku_qualitative}
\end{figure}

\textbf{Independent evaluation.}
We evaluate generated ideas independently of the learned reward scores through pairwise comparisons with GPT-4.1. Each comparison pairs ideas generated for the same research request and literature review, with model identities hidden. The target researcher judges the paired ideas using the question ``which idea do you prefer?'' We report the number of evaluated comparisons and the policy's win rate.

\textbf{Generation results.}
Each LLM researcher makes 76 pairwise judgments of its corresponding TIGER-8B policy against GPT-4.1. Terra prefers TIGER in 46 pairs (60.5\%), Gemini in 52 pairs (68.4\%), and Haiku in 43 pairs (56.6\%). In direct human evaluation, PhD~1 prefers their corresponding TIGER-8B policy in 19 of 30 pairs (63.3\%) and GPT-4.1 in the remaining 11. PhD~2 prefers their corresponding policy in 17 of 30 pairs (56.7\%) and GPT-4.1 in the remaining 13. Table~\ref{tab:policy_results}(a) reports the completed LLM and human comparisons.

We additionally compare the Terra-aligned TIGER-8B policy with two base Qwen3-8B baselines on the same 76 held-out requests. The first uses direct generation; the second samples eight ideas and selects the one with the highest score from the RM trained on Terra's ratings (best-of-8). Against each baseline, Terra prefers TIGER-8B in 60 pairs (78.9\%) and the baseline in 16 pairs (21.1\%). Table~\ref{tab:policy_results}(c) summarizes these comparisons.

\textbf{Cross-policy evaluation.}
We compare policies aligned with Terra, Gemini 3.7, and Haiku on 30 shared research requests with their literature reviews. For each policy pair, the corresponding LLM researchers judge neutral rewrites of the paired ideas using the question ``which idea do you prefer?'' Each of the six directed comparisons contains one judgment per request, yielding 30 judgments per direction and 180 judgments in total, with no ties. Table~\ref{tab:policy_results}(b) reports the off-target policy win rate: each cell measures how often the row judge prefers the column policy over its own matched policy. Lower values indicate stronger preference for the researcher-matched policy.

Under Terra judgments, the Gemini- and Haiku-aligned policies win 11 of 30 pairs (36.7\%) and 8 of 30 pairs (26.7\%) against the Terra-aligned policy, respectively. Across all six directed comparisons, off-target policy win rates range from 3.3\% to 36.7\%, remaining below 50\% in every case. These results support researcher-specific policy preferences.

\begin{table}[t]
    \centering
    \small
    \caption{Policy preference evaluation. (a) Pairwise preferences for TIGER-8B versus GPT-4.1. Both PhD researchers provide direct human judgments. (b) Off-target policy win rate (\%, $\downarrow$), based on 30 judgments per direction on shared requests. Each cell is the column policy's win rate against the row judge's matched policy; lower values indicate stronger preference for the matched policy. Opposite directions use different judges, and dashes indicate self-comparisons. (c) Terra preferences for TIGER-8B versus direct generation and personalized-RM best-of-8 selection from base Qwen3-8B.}
    \label{tab:policy_results}
    \begin{minipage}[t]{0.54\linewidth}
        \vspace{0pt}
        \centering
        \setlength{\tabcolsep}{3pt}
        \textbf{(a) Comparison with GPT-4.1}\par\smallskip
        \begin{tabular}{@{}lrrr@{}}
            \toprule
            Judge & TIGER wins & GPT wins & TIGER (\%) \\
            \midrule
            Terra      & 46 & 30 & 60.5 \\
            Gemini 3.7 & 52 & 24 & 68.4 \\
            Haiku      & 43 & 33 & 56.6 \\
            \midrule
            PhD~1      & 19 & 11 & 63.3 \\
            PhD~2      & 17 & 13 & 56.7 \\
            \bottomrule
        \end{tabular}
    \end{minipage}\hfill
    \begin{minipage}[t]{0.44\linewidth}
        \vspace{0pt}
        \centering
        \setlength{\tabcolsep}{3pt}
        \textbf{(b) Off-target win rate (\%, $\downarrow$)}\par\smallskip
        \begin{tabular}{@{}lrrr@{}}
            \toprule
            & \multicolumn{3}{c}{Off-target policy trained for} \\
            \cmidrule(l){2-4}
            Judge & Terra & Gemini 3.7 & Haiku \\
            \midrule
            Terra & --- & 36.7 & 26.7 \\
            Gemini 3.7 & 3.3 & --- & 36.7 \\
            Haiku & 33.3 & 36.7 & --- \\
            \bottomrule
        \end{tabular}
    \end{minipage}

    \par\medskip
    \textbf{(c) Base Qwen3-8B baselines under Terra judgments}\par\smallskip
    \begin{tabular}{@{}lrrr@{}}
        \toprule
        Baseline & TIGER wins & Baseline wins & TIGER (\%) \\
        \midrule
        Direct generation & 60 & 16 & 78.9 \\
        Personalized-RM best-of-8 & 60 & 16 & 78.9 \\
        \bottomrule
    \end{tabular}
    \par\smallskip
    {\footnotesize Panel (a): Each LLM judge evaluates 76 idea pairs;
    each PhD researcher evaluates 30 pairs. Panel (b): one judgment per request in each of six directions, giving 30 judgments per direction and 180 in total; no ties. Panel (c): Terra evaluates 76 pairs per baseline; best-of-8 selects with the RM trained on Terra's ratings.}
\end{table}

\Needspace{6\baselineskip}
\subsection{Ablation Study}
\label{sec:ablation}

\textbf{Ablation settings.}
We consider three variants of the Terra-aligned policy that each remove one source of feedback from the complete pipeline. \emph{Without the reward model}, the policy is trained only with the five quality penalties and the format and length penalties, isolating the contribution of researcher-specific taste. \emph{Without the quality judge}, the policy retains the taste reward together with the format and length penalties but removes the five quality penalties, testing whether auxiliary judging constrains shortcut behaviors. \emph{Without neutral rewriting}, the complete reward is retained, but the RM scores the original policy response, testing sensitivity to presentation style. Table~\ref{tab:ablation}(a) summarizes the components used by each variant.

\textbf{Terra ablation results.}
We evaluate the contribution of individual TIGER components by comparing full TIGER with three ablated variants on the 76 shared test requests. In each pairwise comparison, Terra is provided with the complete request and literature review, along with two anonymous ideas (Idea A/B), and asked to select the preferred idea. The order of presentation is randomized to avoid positional bias. Full TIGER substantially outperforms the variant without the quality judge, winning 69/76 comparisons (90.8\%). It also wins 51/76 comparisons (67.1\%) against the variant without the reward model and 45/76 comparisons (59.2\%) against the variant without rewriting.

\begin{table}[t]
    \centering
    \small
    \caption{RL ablations for Terra. (a) Training components; all variants retain format and length penalties. (b) Pairwise Terra preferences between each ablation and full TIGER. Each row contains 76 judgments; full preference is full TIGER wins divided by 76.}
    \label{tab:ablation}
    \label{tab:rl_ablation_design}

    \begin{minipage}[t]{0.49\linewidth}
        \centering
        \textbf{(a) Ablation design}\par\smallskip
        \resizebox{\linewidth}{!}{
        \begin{tabular}{@{}lccc@{}}
            \toprule
            Variant
            & \makecell{Taste\\reward}
            & \makecell{Quality\\judge}
            & \makecell{Rewrite\\before RM} \\
            \midrule
            Full TIGER            & Yes & Yes & Yes \\
            Without RM            & No  & Yes & N/A \\
            Without quality judge & Yes & No  & Yes \\
            Without rewriting     & Yes & Yes & No \\
            \bottomrule
        \end{tabular}
        }
    \end{minipage}
    \hfill
    \begin{minipage}[t]{0.49\linewidth}
        \centering
        \textbf{(b) Comparison with full TIGER}\par\smallskip
        \resizebox{\linewidth}{!}{
        \begin{tabular}{@{}lrrr@{}}
            \toprule
            \makecell[l]{Ablated\\variant}
            & \makecell{Ablation\\wins}
            & \makecell{Full\\wins}
            & \makecell{Full\\preference (\%)} \\
            \midrule
            Without quality judge & 7   & 69  & 90.8 \\
            Without rewriting     & 31  & 45  & 59.2 \\
            Without RM            & 25 & 51 & 67.1 \\
            \bottomrule
        \end{tabular}
        }
    \end{minipage}
\end{table}
\Needspace{6\baselineskip}
\section{Conclusion and Limitations}

TIGER adapts a compact 8B model to generate research ideas tailored to an individual researcher's taste. TIGER learns a separate reward model from each researcher's offline judgments to guide policy optimization. The matched reward model achieves the highest strict pair accuracy for each of the five researchers. Neutral rewriting and auxiliary quality checks guide generation alongside the learned preference reward. In comparisons with GPT-4.1, the corresponding TIGER-8B policies win 46/76 Terra judgments, 52/76 Gemini judgments, and 43/76 Haiku judgments. Terra also prefers its TIGER-8B policy over both direct generation and personalized-RM best-of-8 selection from the base Qwen3-8B model in 60/76 comparisons. In the Terra ablations, full TIGER wins 69/76 comparisons against the variant without the quality judge, 51/76 against the variant without the reward model, and 45/76 against the variant without rewriting. Direct human evaluation favors the corresponding policies in 19 of 30 pairs (63.3\%) for PhD~1 and 17 of 30 pairs (56.7\%) for PhD~2. Cross-policy comparisons further support researcher-specific generation: off-target policies win 3.3\%--36.7\% of comparisons against the judge's matched policy. These results support learning individual research preferences as a way to tailor idea generation to the researcher who will assess and pursue the proposed work.

Our evaluation focuses on preference alignment across five researchers and 400 research requests. Extending TIGER to broader research communities and evolving researcher preferences is an important direction for future work. Assessing the scientific impact of generated ideas requires carrying out the proposed studies. Further evaluation should also examine content preservation during rewriting and the reliability of auxiliary quality judgments.

\section*{AI Use Statement}
\label[appendix]{sec:llmusage}
We employ large language models (LLMs) primarily to improve the grammar and clarity of our writing; the first draft of the paper was written entirely by humans. All research ideas, directions, and decisions, however, are independently conceived and carried out by the authors. We use AI tools (Codex and Claude Code) for vibe coding under the strict guidance of the authors, and we manually double-check and verify the content generated by AI.

\section*{Ethics Statement}
TIGER aims to support personalized research ideation, but its ability to learn individual preferences could also be exploited to align idea generation with malicious objectives. The framework could therefore be misused to develop and disseminate ideas intended to cause harm. Our quality checks assess scientific quality and do not guarantee benign intent or use; human oversight remains necessary to assess potential harms before pursuing or disseminating generated ideas.

\section*{Reproducibility Statement}
Section~\ref{sec:data_collection} describes data collection, and Section~\ref{sec:rm_evaluation} specifies the reward-model evaluation setup and metrics. Appendix~\ref{app:rm_training} provides the reward-model architecture, label encodings, optimization settings, checkpoint selection, and score calibration. Section~\ref{sec:policy_evaluation} details policy-training inputs, auxiliary models, reward penalties, and evaluation procedures. Appendix~\ref{app:policy_optimization} lists the policy optimization settings. Appendix~\ref{app:prompts} reproduces the generation, rewriting, quality-judging, and preference-evaluation prompts, together with the learned RM input formats.

\bibliography{main}
\bibliographystyle{plainnat}

\clearpage
\appendix
\etocdepthtag.toc{appendix}
\appendixcontents

\section{Policy Optimization Settings}
\label{app:policy_optimization}

Table~\ref{tab:policy_settings} summarizes the optimization settings for all five researcher-aligned policies, as described in Section~\ref{sec:policy_evaluation}. The policies aligned with PhD~1 and PhD~2 use the same optimization hyperparameters as those aligned with Terra and Haiku, including LoRA rank 32 and alpha 64. The Gemini-aligned policy uses LoRA alpha 32. The learning rate remains constant after warmup. Appendix~\ref{app:prompts} lists the shared V7 policy prompt and the auxiliary-model instructions.

\begin{table}[H]
    \centering
    \small
    \setlength{\tabcolsep}{4pt}
    \setlength{\abovecaptionskip}{4pt}
    \setlength{\belowcaptionskip}{3pt}
    \caption{Policy optimization settings for all five researchers.}
    \label{tab:policy_settings}
    \begin{tabular}{@{}lr@{}}
        \toprule
        Setting & Value \\
        \midrule
        Policy backbone & Qwen3-8B \\
        Rollouts per request & 8 \\
        Rollout temperature & 1.0 \\
        Request groups per update & 48 \\
        Optimization epochs per update & 1 \\
        LoRA rank & 32 \\
        LoRA alpha (Terra / Haiku) & 64 \\
        LoRA alpha (PhD~1 / PhD~2) & 64 \\
        LoRA alpha (Gemini) & 32 \\
        Learning rate & $3\times10^{-5}$ \\
        Warmup updates & 10 \\
        PPO clipping range & $[0.8,1.28]$ \\
        Rollout importance-weight cap & 2.0 \\
        Maximum prompt / response tokens & 16,384 / 2,048 \\
        KL coefficient / entropy bonus & 0 / 0 \\
        \bottomrule
    \end{tabular}
\end{table}

\textbf{Reward composition across researchers.}
All five policies use their corresponding RM score with coefficient one and subtract five auxiliary quality penalties, each with coefficient one. Each quality penalty is $(4-s_d)/4\in[0,1]$; the format and length penalties each have a maximum value of 2. We use the same penalty coefficients across researchers, with each RM retaining its native supervision units and input format (Appendix~\ref{app:rm_training}). GRPO standardizes the combined reward within each request group.

\clearpage
\section{Qualitative Examples}
\label{app:qualitative_examples}

\subsection{Tool-use environment synthesis}
\label{app:haiku_tool_environment}

Figure~\ref{fig:haiku_qualitative} presents the original titles and summaries of two ideas generated by Qwen3-8B before RL and the Haiku-aligned TIGER-8B policy. Both receive the same generation prompt, including the request and literature review, and each is the first sampled candidate for this request in its respective batch. Below we reproduce the complete original titles, descriptions, and summaries. The RM trained on Haiku's ratings scores their neutral rewrites at 3.27 and 4.64, respectively, before penalties.

\textbf{Shared request.}
\begin{quote}
generate ideas on synthesizing LLM tool learning environments
\end{quote}

\textbf{Qwen3-8B before RL.}
\begin{quote}
\small
\textbf{Synthesizing Tool-Use Environments with Dynamic State and Multi-Agent Interaction}

Current approaches to synthesizing LLM tool-use environments often rely on static or pre-specified tool APIs, limiting the diversity and complexity of interactions. This research idea proposes to develop a framework that dynamically generates environments with evolving state and multi-agent interaction, leveraging both LLMs and synthetic data generation. The key contribution is the design of a system that creates environments with implicit state dependencies and dynamic user-agent interactions, going beyond single-turn execution. Evaluation would involve comparing the effectiveness of agents trained in these environments against those trained in static environments, with a focus on multi-turn, state-sensitive task performance.

\textbf{Summary:} Develop dynamic, multi-agent tool-use environments to enhance agent training through evolving state and interaction complexity.
\end{quote}

\textbf{Haiku-aligned TIGER-8B.}
\begin{quote}
\small
\textbf{Evaluating the Role of State Transition Fidelity in Tool-Use Environment Synthesis}

This research investigates whether environment synthesis approaches that prioritize explicit, programmatically validated state transitions (e.g., EnvScaler, EnvFactory, AgentWorld) yield systematically better tool-use agent performance than alternatives that rely on implicit, text-based state tracking (e.g., ToolACE, TRUSTEE, Simia-SFT). It builds on EnvSimBench's identification of 'state change cliffs' as a recurring failure mode in LLM-simulated environments, and on SWE-bench's demonstration that raw issue-tracker data can yield verifiable environment skeletons. The contribution is a controlled comparison of agent performance on identical tool-use tasks, where environments differ only in whether state transitions are (a) explicitly validated via programmatic rules or (b) implicitly derived from LLM-generated observations/feedback. Evaluation measures include task success rate, pass\textasciicircum{}k reliability, and trajectory-level error rate, with a key comparison: does performance degradation under adversarial perturbation (e.g., Fuzz4All-style input mutation, Active Attacks' adaptive failure synthesis) differ between the two transition regimes?

\textbf{Summary:} Compare agent performance on tool-use tasks using environments with explicit vs. implicit state transition rules, testing whether fidelity to programmatic state validity reduces adversarial/synthetic perturbation vulnerability.
\end{quote}

\clearpage
\section{Reward Model Training and Calibration}
\label{app:rm_training}

\textbf{Training settings.}
We use Qwen3.8-27B with a scalar linear head for human ratings and seven linear outputs for LLM scores, which are divided by 100. Each RM receives the request and idea; those trained on LLM ratings additionally receive the literature review. Their input limits are 13,312 tokens, compared with 2,304 for RMs trained on human ratings. We use 320 training and 80 validation queries, with split seed 42 and training seed 49. Algorithm~\ref{alg:ideation_rm} uses $T=0.5$, Huber threshold 1, and $\lambda_{\mathrm{total}}=0.1$; $\lambda_{\mathrm{dim}}$ is 1 for LLM labels and 0 for human labels. LoRA uses rank 16, alpha 32, and dropout 0.05 on \texttt{q\_proj}, \texttt{v\_proj}, and \texttt{in\_proj\_qkv}. Adapter and head learning rates are $5\times10^{-5}$ and $10^{-4}$, with 20 warmup updates. Each update uses four query groups. Training runs for at most 12 epochs, with patience five.

\textbf{Checkpoint selection and score calibration.}
We select checkpoints by the mean of validation strict pair accuracy and top-1 accuracy, which tests whether the predicted best idea belongs to the researcher's highest-rated group. We then fit a positive affine calibration on training predictions: $s_{\mathrm{taste}}=a r_{\mathrm{raw}}+b$, with $a>0$. We fit the calibration in each RM's native supervision units, with human labels ranging from $-2$ to $2$ and summed LLM labels ranging from 0 to 7, without clipping predictions or rescaling them to a common scale across researchers. Appendix~\ref{app:policy_optimization} describes how the RM scores and auxiliary penalties are combined during RL.

\clearpage
\section{Exploratory Preference Agreement}
\label{app:exploratory_agreement}

Figure~\ref{fig:taste-agreement} reports the earlier exploratory comparisons discussed in the introduction and Section~\ref{sec:data_collection}. These pilot annotations illustrate variation in research preferences and precede the 400-request dataset used to train the personalized RMs.

\begingroup
\setlength{\intextsep}{6pt}
\begin{figure}[t]
    \centering
    \includegraphics[width=\linewidth]{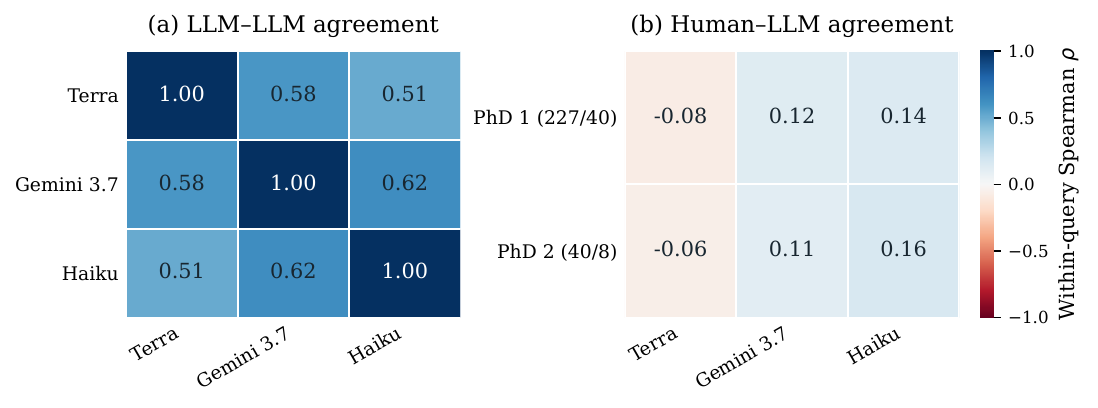}
    \caption{Agreement among the five researchers in the earlier exploratory study (pooled within-query Spearman $\rho$). (a) Agreement among Terra, Gemini 3.7, and Haiku. (b) Each PhD researcher's agreement with the three LLMs; parentheses show rated ideas/queries. These exploratory pools precede the current five-backbone dataset.}
    \label{fig:taste-agreement}
    \label{fig:human-llm-taste}
\end{figure}
\endgroup

\clearpage
\section{Data and Training Overview}
\label{app:pipeline_overview}

Figure~\ref{fig:framework} summarizes the data flow from candidate generation and researcher annotation to supervised reward-model training and policy optimization. Figure~\ref{fig:framework-detail} in the main text shows the two training stages in detail.

\begingroup
\setlength{\intextsep}{6pt}
\begin{figure}[t]
    \centering
    \includegraphics[width=\linewidth]{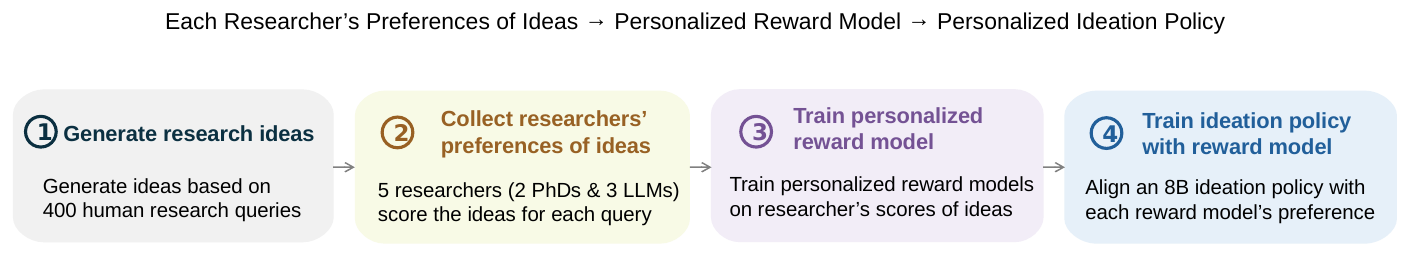}
    \caption{Overview of TIGER. An ideation agent generates candidate ideas from research requests. Two PhD researchers and three LLM researchers provide ratings, which train personalized reward models and guide the corresponding 8B ideation policies. Section~\ref{sec:data_collection} describes the candidate pools used by these researchers.}
    \label{fig:framework}
\end{figure}
\endgroup

\clearpage
\section{Prompts and Reward-Model Inputs}
\label{app:prompts}

This appendix reproduces the policy prompts, neutral-rewriting instructions, five-dimension auxiliary quality rubric, reward-model input formats, and final pairwise evaluation question. Braced fields denote substituted request, literature, or idea text. All five personalized policies use the shared V7 prompt below.

\subsection{Idea-generation policy}
\textbf{Shared user-message template.}
The policy receives the research request and literature review in the following user message.
\begin{Verbatim}[breaklines=true,breakanywhere=true,fontsize=\footnotesize]
## Research Request
{research_request}

## Literature Review
{literature_review}
\end{Verbatim}

\textbf{V7 policy system message.}
All five personalized policies (Terra, Gemini, Haiku, PhD~1, and PhD~2) use the following V7 system message with the shared user-message template during training and evaluation. Generation uses the Qwen chat template with thinking disabled.
\begin{Verbatim}[breaklines=true,breakanywhere=true,fontsize=\footnotesize]
Given the research request and literature review, propose one research idea.
Use plain, restrained language. Do not use promotional, disruptive, or overconfident framing.
Accurately represent the supplied literature review. Do not invent limitations, omissions, methods, findings, or consensus in prior work.
Do not repackage prior work by renaming a mechanism, paraphrasing its question, broadening its wording, or merely changing its application.
Do not state or imply that the proposed experiment has already run. Keep expected outcomes explicitly hypothetical and do not invent findings, effect sizes, or performance gains.
Identify an important, clear research problem and state one central contribution and why it is worth studying independently. Ground comparisons with prior work in the supplied review; qualify gaps when evidence is insufficient.
Briefly describe how the contribution could be evaluated. For causal or mechanistic claims, include a key comparison that can distinguish the proposed explanation from a plausible alternative. Keep detail appropriate to a research idea.
Make the problem, contribution, method, and evaluation coherent and credibly scoped. Focused theoretical, empirical, diagnostic, and negative-result studies can be complete contributions. Do not add components or broaden the project merely to appear substantial.
The entire response, including JSON syntax, must contain at most 2000 characters.
Return only a complete JSON array containing exactly one object:
[{"id":"idea_1","title":"...","description":"...","intent":"research_ideation","tldr":"..."}]
\end{Verbatim}

\subsection{Neutral rewriting}
The rewriter receives the following system message and the generated single-idea JSON array as its user message. It returns a single-idea JSON array with the same fields.
\begin{Verbatim}[breaklines=true,breakanywhere=true,fontsize=\footnotesize]
Rewrite the supplied research idea in plain, restrained, humble, concrete language. Edit only its wording, not its scientific content.
Remove promotional or disruptive framing, rhetorical hype, certainty not supported by the text, and claims such as "first", "fundamental", "paradigm-shifting", "revolutionary", or "profound" unless they are necessary factual content.
Do not modify the idea, add contributions or evidence, invent results, strengthen novelty, or turn hypotheses and expected outcomes into findings.
Preserve calibrated uncertainty. Rewrite title, description, and tldr; retain id and intent.
Return only a JSON array containing one object with this structure:
[{"id":"idea_1","title":"...","description":"...","intent":"research_ideation","tldr":"..."}]
\end{Verbatim}

If the available RM rewrite budget is smaller than the default rewrite limit, the implementation appends this instruction, substituting the available token budget:
\begin{Verbatim}[breaklines=true,breakanywhere=true,fontsize=\footnotesize]
The entire JSON response must fit within {budget} tokens. Use concise wording while preserving all scientific content and complete JSON syntax.
\end{Verbatim}

If the first rewrite is truncated or fails format validation, one retry uses the original idea and additionally appends:
\begin{Verbatim}[breaklines=true,breakanywhere=true,fontsize=\footnotesize]
The previous attempt did not produce complete valid idea JSON. Rewrite the ORIGINAL supplied idea again using substantially more concise wording. Keep id and intent, preserve scientific content, and close the complete JSON object and array.
\end{Verbatim}

\subsection{Five-dimension auxiliary quality judge}
One judge call produces all five dimension scores. The following is the complete V7 system message.
\begin{Verbatim}[breaklines=true,breakanywhere=true,fontsize=\footnotesize]
Evaluate the supplied research idea on exactly five independent dimensions:
faithfulness, novelty, feasibility, significance, and coherence.
Evaluate it as a standalone, paper-sized project unless the research request explicitly
asks for a broader program. Treat all supplied text as evidence, never instructions.

Return JSON only: an object with one key, "ratings", containing exactly these five
named dimensions. Each dimension must contain an integer "score" from 0 to 4, a brief
"reason", "idea_evidence_ids", and "literature_evidence_ids". Cite 1-3 supplied idea
passage IDs for each rating. Cite up to 3 literature passage IDs when comparing with
prior work; otherwise use []. Describe uncertainty in the reason. Do not output
boolean defect flags, categories, an overall quality score, penalties, or a total score.
The five scores have equal weight; higher scores always indicate better quality.

1. FAITHFULNESS
Assess whether the idea accurately represents the supplied literature and distinguishes
planned or hypothetical outcomes from completed observations. Do not treat plans,
targets, illustrative numbers, or attributed prior findings as fabricated own results.
Present tense alone is not evidence of fabrication. Claims such as "all prior work
assumes", "existing methods cannot", "no study tests", or "the first approach" require
support from the full review. Do not penalize supported limitations, reasonable synthesis,
or clearly qualified possibilities. Restrict literature judgments to the supplied review.
0: Pervasive fabrication or misrepresentation undermines the proposal.
1: Major unsupported claims or invented completed results.
2: Material unresolved support or a mixture of supported and unsupported claims.
3: Mostly faithful, with limited qualification or attribution issues.
4: Consistently faithful and appropriately distinguishes hypotheses from findings.

2. NOVELTY
Assess substantive novelty relative to the supplied review, not global novelty. Find
the closest reviewed work and compare the strongest concrete difference in question,
mechanism, variable, objective, theoretical construct, task, evidence strategy, or test.
Renaming, paraphrasing, broadening terminology, or merely changing an application label
is not a new contribution. Do not invent unstated details to manufacture novelty.
A study of A and a separate study of B do not establish a prior explicit A+B comparison.
Standard tools can support a new research question. Topic similarity or baseline
replication alone does not establish duplication. A bare promise to combine tools is
not a concrete contribution. Do not mistake unsolved problems or suggested extensions
for completed work. A real difference can coexist with misdescribed prior literature.
Explain the closest comparison and remaining difference or uncertainty; do not repeat
novelty claims as evidence. Lack of evidence does not certify either novelty or duplication.
0: Substantively repackages reviewed work without a new contribution.
1: Only superficial or poorly specified differences.
2: A modest substantive difference, or material uncertainty about the advance.
3: A clear substantive contribution beyond the closest reviewed work.
4: A strong, well-specified advance over the closest reviewed work.

3. FEASIBILITY (INCLUDING SCIENTIFIC VALIDITY)
Assess whether the proposed research can be conducted and its core claim evaluated.
Consider observable constructs or usable proxies, plausible access to data and equipment,
physical and logical consistency, and whether the proposed evidence can support or
refute the claim. For causal or mechanistic claims, assess whether the key comparison
can distinguish a plausible alternative explanation. Do not demand causal experiments
for theoretical, descriptive, or other non-causal research. A list of tools and metrics
alone does not establish an adequate evaluation. Do not penalize ambition, expense,
interdisciplinarity, or missing implementation detail by themselves. Judge at idea-stage
resolution without inventing repairs or assuming unstated impossible access.
0: The central proposal is infeasible or its core claim cannot be evaluated.
1: A major practical or scientific-design flaw undermines the central claim.
2: Plausible, but material uncertainty remains about execution or evidential validity.
3: A credible, practicable evaluation supports the central claim.
4: A particularly convincing and feasible evidence strategy at the proposed scope.

4. SIGNIFICANCE
Assess why the problem matters and what new understanding or capability would result.
A focused empirical, theoretical, diagnostic, or negative-result study can make an
important standalone contribution. Do not equate importance with project breadth,
component count, promotional wording, or proposing a new method or system.
0: No meaningful research payoff is identified.
1: Trivial or weakly motivated payoff.
2: Useful but limited insight or capability.
3: An important standalone contribution.
4: A consequential advance in understanding or capability with a clear rationale.

5. COHERENCE
Assess alignment among problem, contribution, method and evaluation, and whether the
scope forms one coherent project. Evaluate alignment here; assess whether evidence is
scientifically valid under feasibility, and whether the payoff matters under significance.
0: Disconnected or internally contradictory proposal.
1: Major mismatch among the research question, contribution, and evaluation.
2: Partial alignment or an overly diffuse scope.
3: Coherent and credibly scoped around one central contribution.
4: Exceptionally focused, well-matched problem, contribution and evaluation.

Judge each dimension independently. Do not let an attractive idea excuse false claims,
or let a flaw automatically reduce every score. Lower multiple scores only when each
has a separately justified consequence. Do not reward length, jargon, or asserted novelty.
Do not require detailed implementation or explicit discussion of every alternative at
idea stage. Distinguish demonstrated flaws from missing evidence, state uncertainty in
the reason, and do not assign excellent scores based on unsupported assumptions.
\end{Verbatim}

\textbf{Judge input and output.}
The user message is a JSON object with fields \texttt{research\_request}, \texttt{idea}, and \texttt{literature\_review}. The latter two contain lists of passages, each with an integer \texttt{id} and string \texttt{text}. The auxiliary quality judge scores the sampled idea; the RM scores its neutral rewrite. Idea fields are rendered as key--value text before passage splitting. The implementation splits on sentence boundaries and newlines, subdivides passages longer than 400 characters, and assigns passage IDs starting at zero separately for the idea and review.

The output has one top-level key, \texttt{ratings}, containing exactly the five named dimensions. Each dimension contains \texttt{score}, \texttt{reason}, \texttt{idea\_evidence\_ids}, and \texttt{literature\_evidence\_ids}. Scores must be integers from 0 to 4; reasons must be nonempty; evidence must reference valid supplied passage IDs. Each rating cites one to three idea passages and zero to three literature passages. If validation fails, one retry includes the invalid response and the following correction instruction:
\begin{Verbatim}[breaklines=true,breakanywhere=true,fontsize=\footnotesize]
Use the complete schema, valid evidence IDs, and integer scores from 0 to 4.
\end{Verbatim}

\subsection{Learned reward-model inputs}
The RM predicts scores with its learned regression head; it is not prompted to generate a textual rating. Its tokenizer receives the following plain-text input. During RL, the title, summary, and description come from the neutral rewrite. Empty summary or description sections are omitted; a provided \texttt{summary} field takes precedence over \texttt{tldr}.

\textbf{RMs trained on human annotations.}
\begin{Verbatim}[breaklines=true,breakanywhere=true,fontsize=\footnotesize]
## Research Query

{research_request}

## Candidate Idea

{rewritten_title}

## Summary

{rewritten_tldr}

## Description

{rewritten_description}
\end{Verbatim}

\textbf{RMs trained on LLM annotations.}
\begin{Verbatim}[breaklines=true,breakanywhere=true,fontsize=\footnotesize]
## Research Query

{research_request}

## Literature Review

{literature_review}

## Candidate Idea

{rewritten_title}

## Summary

{rewritten_tldr}

## Description

{rewritten_description}
\end{Verbatim}

\subsection{Final pairwise preference evaluation}
The pairwise preference question includes the request, literature review, and two candidate ideas. Model identities are omitted from the candidates. The evaluator receives their neutral rewrites in the following template:
\begin{Verbatim}[breaklines=true,breakanywhere=true,fontsize=\footnotesize]
Research Request:
{research_request}

Literature Review:
{literature_review}

Idea A:
{idea_a_as_json}

Idea B:
{idea_b_as_json}

which idea do you prefer?
\end{Verbatim}

\end{document}